\def\paperlanguage{}
\pdfoutput=1 

\documentclass[letterpaper, 10 pt, conference]{ieeeconf}  

\usepackage{bm}
\usepackage{cite}
\usepackage{flushend}
\include{preamble}

\IEEEoverridecommandlockouts                              

\title{\LARGE \textbf
  {
    \switchlanguage%
    {%
      Stability Recognition with Active Vibration for Bracing Behaviors and Motion Extensions Using Environment in Musculoskeletal Humanoids
    }%
    {%
      筋骨格ヒューマノイドのBracingと動作拡張のための能動的振動を伴う安定性認識
    }%
  }
}

\author{Kento Kawaharazuka$^{1}$, Manabu Nishiura$^{1}$, Shinsuke Nakashima$^{1}$, Yasunori Toshimitsu$^{1}$, Yusuke Omura$^{1}$\\Yuya Koga$^{1}$, Yuki Asano$^{1}$, Kei Okada$^{1}$, Koji Kawasaki$^{2}$, and Masayuki Inaba$^{1}$
  \thanks{$^{1}$ The authors are with the Department of Mechano-Informatics, Graduate School of Information Science and Technology, The University of Tokyo, 7-3-1 Hongo, Bunkyo-ku, Tokyo, 113-8656, Japan.
    {\texttt\small [kawaharazuka, nishiura, snakashima, toshimitsu, oomura, koga, asano, k-okada, inaba]@jsk.t.u-tokyo.ac.jp}
  }
  \thanks{$^{2}$ The author is associated with TOYOTA MOTOR CORPORATION.
    {\texttt\small koji\_kawasaki@mail.toyota.co.jp}
  }
}

\begin{document}

\maketitle
\thispagestyle{empty}
\pagestyle{empty}

\begin{abstract}
  \switchlanguage%
  {%
    Although robots with flexible bodies are superior in terms of the contact and adaptability, it is difficult to control them precisely.
    On the other hand, human beings make use of the surrounding environments to stabilize their bodies and control their movements.
    In this study, we propose a method for the bracing motion and extension of the range of motion using the environment for the musculoskeletal humanoid.
    Here, it is necessary to recognize the stability of the body when contacting the environment, and we develop a method to measure it by using the change in sensor values of the body when actively vibrating a part of the body.
    Experiments are conducted using the musculoskeletal humanoid Musashi, and the effectiveness of this method is confirmed.
  }%
  {%
    柔軟な身体を持つロボットは接触や適応性の観点から優れるものの, 正確な制御が難しい.
    これに対して, 人間は周囲の環境を利用することで身体を安定化し, 手先のブレを抑制する.
    本研究では, 筋骨格ヒューマノイドを題材として, 環境を使ったbracing, 動作範囲の拡張を行う.
    この時, 環境に触れた際の身体の安定化を認識する必要があり, 身体の一部を能動的に振動させた際の身体のセンサ値の変化を利用してこれを測定する手法を開発した.
    筋骨格ヒューマノイドMusashiを用いて実験を行い, その有効性を確認した.
  }%
\end{abstract}

\section{INTRODUCTION}\label{sec:introduction}
\begin{figure}[t]
  \centering
  \includegraphics[width=0.7\columnwidth]{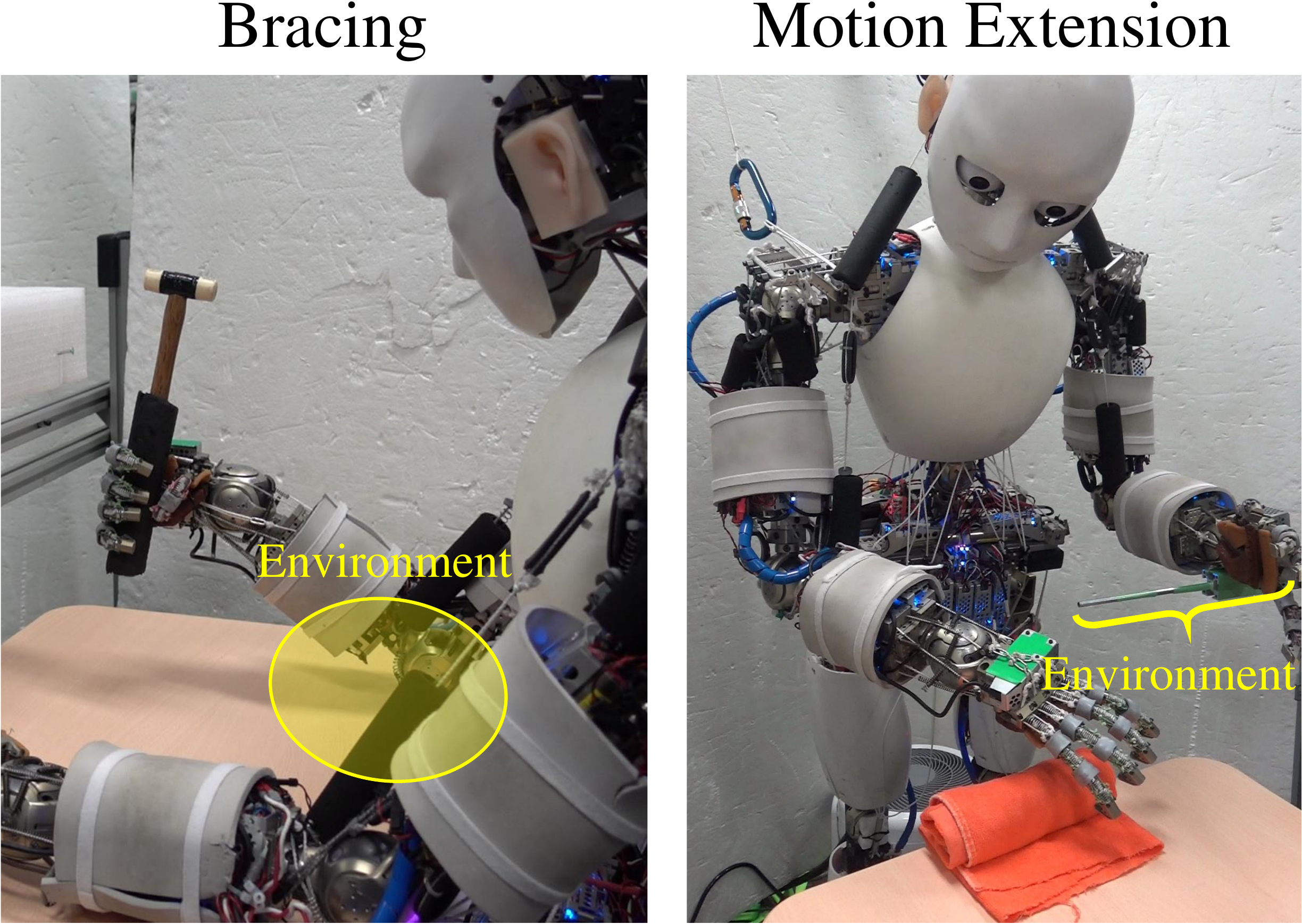}
  \vspace{-1.0ex}
  \caption{Bracing behavior and motion extension using environment.}
  \label{figure:motivation}
  \vspace{-3.0ex}
\end{figure}

\switchlanguage%
{
  The flexible body is excellent from the point of view of the soft contact, impact mitigation, adaptability, etc. \cite{kim2013softrobotics, lee2017softrobotics}, and a shift from rigid robots \cite{kaneko2002hrp, kojima2015jaxon} to soft robots \cite{niiyama2010athlete, nakanishi2013design} is underway.
  In \cite{niiyama2010athlete}, a robot that jumps and runs using pneumatic artificial muscles is developed.
  In \cite{nakanishi2013design}, a robot that mitigates impact and softly interacts with the environment using variable stiffness control with nonlinear elastic elements has been developed.

  However, since the flexible body is difficult to control and move as intended, walking controls or precise movements as are possible with a rigid robot are difficult for such soft robots.
  In order to solve this problem, various control methods have been developed so far.
  In \cite{moudgal1995flexible}, a learning control of a flexible link robot with fuzzy control is developed.
  In \cite{kawaharazuka2019flexible}, the equation of motion including the image of the flexible body is trained and the accurate dynamic motion is realized.
  In \cite{hofer2019iterative}, a soft robotic arm is accurately controlled using the iterative learning control.

  In contrast, humans stabilize their bodies and perform precise movements with the bracing behavior by propping up or leaning on the environment (\figref{figure:motivation}).
  Accurate movement with the forearms, wrists or elbows attached to a desk or other objects is called ``bracing'', and research on robots using this method was started in the 1980s.
  In addition to studies on accurate and energy efficient movement generation with bracing in industrial robots \cite{hollis1992bracing, zupancic1998bracing, li2017bracing}, applications to a dental robotic system \cite{li2019dental} and a wearable device \cite{parietti2014bracing} have been developed.
  It is also possible to extend its own workspace and reach out to more distant locations by balancing the body using the environment.
  In \cite{harada2004environment}, the stability of the body is increased by grasping the environment, and in \cite{khatib2014supported}, the uneven terrain is traversed by increasing contact with the environment through the use of smart staff.
  In \cite{henze2014multicontact}, a multi-contact control method to the environment is proposed using model predictive control.

  However, these methods have been applied only to rigid axis-driven robots and not to flexible robots.
  The reason for this is that it is difficult for a flexible robot to accurately recognize its own posture and make contact with the environment as intended.
  Most previous studies have assumed that the intended posture is accurately realized as is modelized, and some studies have solved this problem by allowing humans to operate the robot.
  Therefore, in order to apply the bracing motion to a flexible robot, it is necessary to change the way of thinking that the robot should not move exactly as intended, but should first try to move and then recognize the stability.
  This perception of stability is key for flexible robots, and the environment cannot be used without it.
  Since it is more difficult for flexible robots to move accurately than rigid robots, the effect of bracing behavior is considered to be significant.
  Also, by using the environment, the robots can extend the range of motion by increasing the stability of its own body.
}%
{%
  柔軟な身体は柔らかい接触や衝撃の緩和, 適応性等の観点から優れており\cite{kim2013softrobotics, lee2017softrobotics}, これまでの固いロボット\cite{kaneko2002hrp, kojima2015jaxon}から柔らかいロボット\cite{niiyama2010athlete, nakanishi2013design}への転換が推し進められている.
  \cite{niiyama2010athlete}では空気圧人工筋肉を用いてジャンプしたり走ったりするロボットが構成されている.
  \cite{nakanishi2013design}では非線形弾性要素を用いた可変剛性により衝撃を緩和したり, 柔らかく環境と触れ合うロボットが構成されている.

  しかしその制御は難しく, 意図したとおりに正確に動くことができないため, 固いロボットで実現されるような歩行制御や, 精密な動きは困難である.
  これを解決するために, これまで様々な制御手法が開発されている.
  \cite{moudgal1995flexible}ではfuzzy制御を用いたflexible link robotの学習型制御が開発されている.
  \cite{kawaharazuka2019flexible}では身体の画像を含む運動方程式を学習し, 正確な動的動作の実現がされている.
  \cite{hofer2019iterative}では, iterative learning controlを用いてsoft robotic armを正確に制御している.

  これらに対して人間は, 環境に手をついたり, 寄っかかったりすることで, 自分の身体を安定化したり, 手先のブレを抑えて正確に動作を行ったりする(\figref{figure:motivation}).
  前腕や手首, 肘を机等につけて正確に動く振る舞いはbracingと呼ばれ, 1980年代にはこれを用いたロボットの研究が始まっている\cite{book1985bracing}.
  industrial robotにおけるbracingの活用による正確で消費エネルギーの少ない動作生成に関する研究\cite{hollis1992bracing, zupancic1998bracing, li2017bracing}の他, dental robotic system \cite{li2019dental} や wearable device \cite{parietti2014bracing}への応用も成されている.
  また, 環境を使ってバランスを取ることで, 自分の作業空間を拡張し, より遠いところへ手を伸ばすことが可能である.
  \cite{harada2004environment}では環境を把持することで安定性を増し, \cite{khatib2014supported}ではsmart staffを用いて環境接触を増やすことでuneven terrainを踏破している.
  \cite{henze2014multicontact}ではmodel predictive controlを用いて環境等に対するmulti-contactな制御法を提案している.

  しかし, これらの手法は軸駆動型の固いロボットにのみ適用されており, 柔軟なロボットへの適用は見られない.
  これは, 柔軟なロボットが正確に自分の姿勢を把握して意図した通りに環境接触を行うことが難しいことが理由として考えられる.
  ほとんどの先行研究はモデル化から正確にその姿勢を実現することを前提としており, 一部の研究では人間がロボットを操作することでこの問題を解決している.
  よって, 柔軟なロボットにbracingを適用するためには, 計画した通りに正確に動くのではなく, とりあえず動いてみてその後安定性を認識する, という発想の転換が必要である.
  この安定性の認識が肝であり, その認識なくして環境を用いることはできない.
  柔軟なロボットは固いロボットよりも正確な動きが難しいため, bracingの効果は大きいと考えられる.
  また, 柔軟ゆえに難しく, あまり扱われることのないバランスについても, 環境を用いることで安定性を増し, より動作範囲を広げることができると考えられる.
}%

\switchlanguage%
{%
  The objective of this study is to extend the range of motion and reduce the shake of the flexible musculoskeletal humanoid \cite{nakanishi2013design, kawaharazuka2019musashi} by using the environment to stabilize the body.
  As mentioned above, the most important issue for a flexible robot is the perception of stability, and we focus on its development.
  In this study, we consider that the stability can be measured by the degree of suppression of the vibration when a part of the body is actively vibrated.
  While vibrations have been suppressed so far, we dared to generate vibrations and use them for stability recognition.
  We use this to monitor the stability of the body, let the body come into contact with the environment, and then perform some movements after stabilizing the body.
  Although this method is applied to the musculoskeletal humanoid in this study, the same concept can be applied to any robot, and we believe that this simple concept will be a new powerful tool for flexible robots.

  The contributions of this study are listed below.
  \begin{itemize}
    \item Development of a stability evaluation method with active vibration
    \item Consideration of the change in evaluated values due to changes in parameters of the active vibration and observation
    \item Experiments on bracing behavior and motion extension in musculoskeletal humanoids using the stability recognition
  \end{itemize}

  The structure of this study is organized as follows.
  First, the overview of the musculoskeletal humanoid, the method of active vibration, and the method of observing the vibration of the body are briefly described.
  Second, various preliminary experiments are conducted to investigate the differences in the stability recognition with different parameters of the vibration.
  Based on the results, we conduct experiments on motion extension and bracing behavior using the environments.
}%
{%
  本研究は, 人体を模倣した柔軟な筋骨格ヒューマノイド\cite{nakanishi2013design, kawaharazuka2019musashi}を題材として, 環境を使って身体を安定化させ, 動作範囲を拡張したり, ブレを抑えたりすることを目的とする.
  これまでに述べたように, 柔軟なロボットが環境を用いる中で, 安定性の認識が最も重要であり, 本研究ではその開発に主眼を置く.
  本研究ではこの安定性を, 自分の身体に能動的に振動を加えた時, その振動がどの程度抑制されたかによって計測可能であると考えた.
  これまで振動は抑制するものであったが\cite{kiang2015flexible}, 敢えて振動を起こすことで, これを安定性認識に利用する.
  これを用いて身体の安定化度合いをモニタリングし, 身体を環境に接触させ, 安定化させた後に動作を行う.
  非常にシンプルな手法と考え方であるが, 柔軟なロボットにとっては新しい強力なツールになると考える.

  本研究のcontributionを以下に列挙する.
  \begin{itemize}
    \item 能動的振動を用いた安定性評価方法の開発
    \item 能動的振動と観察のパラメータ変化による評価値変化の考察
    \item 安定性評価を用いた筋骨格ヒューマノイドのbracingと動作拡張実験
  \end{itemize}

  本研究の構成は以下のようになっている.
  まず, 筋骨格ヒューマノイドの概要, 自己身体の振動方法とその観測方法について簡潔に述べる.
  次に, 自己身体の振動に関する様々な予備実験を行い, 振動箇所・振動の観測箇所・周波数・振幅の違いによる観測結果の違いについて実験する.
  その結果をもとに, 身体安定化による動作拡張, bracingによる体幹の安定化実験を行う.
}%

\begin{figure}[t]
  \centering
  \includegraphics[width=0.7\columnwidth]{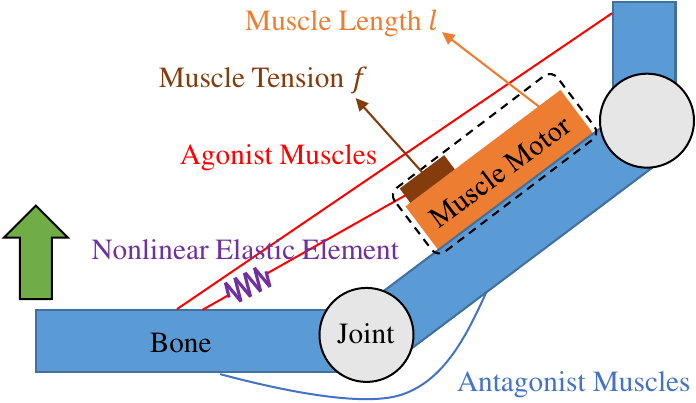}
  \vspace{-1.0ex}
  \caption{The basic musculoskeletal structure.}
  \label{figure:musculoskeletal-structure}
  \vspace{-3.0ex}
\end{figure}

\section{Active Vibration and Stability Recognition for Musculoskeletal Humanoids} \label{sec:proposed-methods}
\subsection{Musculoskeletal Humanoids} \label{subsec:musculoskeletal-humanoids}
\switchlanguage%
{%
  The basic musculoskeletal structure is shown in \figref{figure:musculoskeletal-structure}.
  Redundant muscles are antagonistically arranged around the joint.
  The muscles are mainly composed of Dyneema, an abrasion resistant synthetic fiber, and nonlinear elastic elements enabling variable stiffness control are often arranged in series with the muscles.
  In some robots, the muscles are folded back with pulleys in order to gain the moment arm, and high friction is often generated at their sliding parts.
  For each muscle, muscle length $l$, muscle tension $f$, and muscle temperature $c$ can be measured from the encoder, load cell, and temperature sensor, respectively.
  The joint angle $\bm{\theta}$ is usually difficult to measure due to the ball joint and complex scapula, and the flexible body is difficult to modelize.
  In this study, not muscle tension but muscle length command $l^{ref}$ with high trackability is used as control input in order to vibrate at a fast cycle.

}%
{%
  基本的な筋骨格構造の構成を\figref{figure:musculoskeletal-structure}に示す.
  冗長な筋が関節の周りに拮抗して配置されている.
  筋は主に摩擦に強い合成繊維であるDyneemaによって構成されており, 可変剛性を可能とする非線形性弾性要素が筋と直列に配置されている場合が多い.
  ロボットによっては, モーメントアームを稼ぐために滑車を使って筋を折り返している場合もあり, それら摺動部に大きな摩擦が生じることがしばしばある.
  それぞれの筋についてエンコーダから筋長$l$・ロードセルから筋張力$f$・温度センサから筋温度$c$が測定できる.
  関節角度$\bm{\theta}$は球関節や複雑な肩甲骨ゆえに測定できない場合が多く, モデル化も困難である.
  制御入力として筋張力, 筋長の両者が考えられるが, 本研究では速い周期で振動入力を与えるため, より追従性の速い筋長指令$l^{ref}$を用いている.
}%

\begin{figure}[t]
  \centering
  \includegraphics[width=0.8\columnwidth]{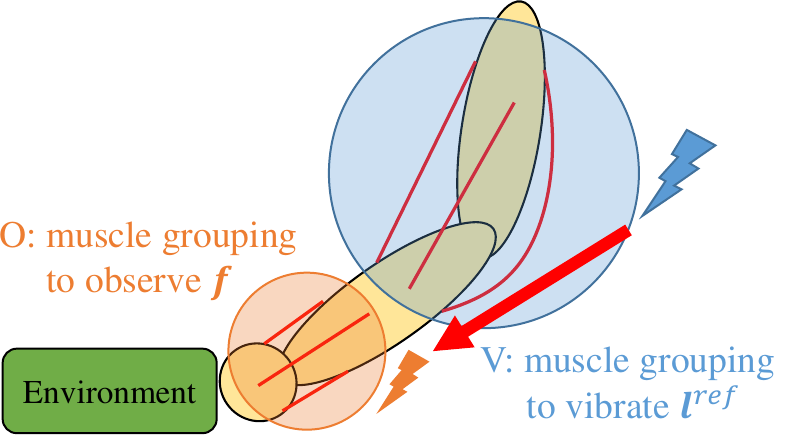}
  \vspace{-1.0ex}
  \caption{The concept of active vibration and stability recognition.}
  \label{figure:active-vibration}
  \vspace{-3.0ex}
\end{figure}

\subsection{Active Vibration and Stability Recognition} \label{subsec:active-vibration}
\switchlanguage%
{%
  The procedure for stability recognition using active vibration is as follows (\figref{figure:active-vibration}).
  \begin{enumerate}
    \item Determine the muscle group $V$ that vibrates and the muscle group $O$ that observes the vibration.
    \item Vibrate the muscle length control command $\bm{l}^{ref}$ of the muscle group $V$.
    \item Detect the degree of vibration propagation using the muscle tension $\bm{f}$ of the muscle group $O$.
  \end{enumerate}

  Regarding 1), since the way to choose $V$ and $O$ varies depending on the robot, we will discuss this in \secref{sec:preliminary-experiments}.
  Let $M_{\{V, O\}}$ be the number of muscles in $V$ or $O$, and $\bm{l}^{ref}_{\{V, O\}}$ be the muscle length command of muscles in $V$ or $O$, and $\bm{f}_{\{V, O\}}$ be the muscle tension of muscles in $V$ or $O$.

  Regarding 2), we update the muscle length command $\bm{l}^{ref}_{V}$ of each muscle in $V$ as shown below,
  \begin{align}
    l^{ref'}_{V} = l^{ref}_{V} + A\textrm{sin}(2{\pi}Ft)
  \end{align}
  where $t$ is the current time, $F$ is the frequency of vibration, $A$ is the amplitude of vibration, $\bm{l}^{ref}_{V}$ is the original muscle length command, and $\bm{l}^{ref'}_{V}$ is the muscle length command sent to the actual robot after adding the vibration value to $\bm{l}^{ref}_{V}$.

  Regarding 3), using the muscle tension $\bm{f}_{O}$ of muscles in $O$, the evaluation value $E$ is calculated as follows,
  \begin{align}
    \bm{h} &= \textrm{Extract}_{[F_{min}, F_{max}]}(\textrm{FFT}(\bm{f}_{O, [t-T+1, t]})) \label{eq:fft}\\
    E_{raw} &= \frac{1}{M_{O'}}\sum_{i \in O'} h_{i} \label{eq:ave}\\
    E &\gets (1-{\alpha})E+{\alpha}E_{raw} \label{eq:lowpass}
  \end{align}
  where $\textrm{FFT}$ is the Fourier transform, $T$ is the data length for $\textrm{FFT}$, $\bm{f}_{O, [t-T+1, t]}$ is the time series data of $\bm{f}_{O}$ within the range of $[t-T+1, t]$, and $\textrm{Extract}_{[F_{ min}, F_{max}]}$ represents a function to extract the largest Fourier transformed value in a specified frequency range $[F_{min}, F_{max}]$.
  That is to say, \equref{eq:fft} is an operation that takes out the largest spectrum in a given frequency range by separately conducting Fourier transform for each muscle in $O$.
  Although it is possible to perform Fourier transform for only the L2 norm of $\bm{f}_{O}$ without separately conducting it for each muscle, this method is used because the phase of the vibrations of observed muscle tensions can be different for each muscle even if the original vibrations of muscles are in the same phase.
  Also, $O'$ is a grouping where two muscles with large values of $h$ and two muscles with small values of $h$ are subtracted from $O$.
  That is, \equref{eq:ave} expresses the average of $h$ of muscles in $O'$.
  This procedure is a heuristic obtained from the fact which we found through experiments: when an external force is applied due to contact with the environment, some muscles are highly loaded and this sometimes can change the degree of propagation of vibrations.
  This procedure enables us to handle the suppression of vibrations by contact with the environment, rather than the load from the environment.
  If the number of muscles is small, it is possible to skip this procedure and take the average of $h$ for all the muscles in $O$, which reduces the performance but does not significantly change the characteristics of the system.
  Finally, we apply a low pass filter \equref{eq:lowpass} with $\alpha$ as a coefficient.
  The change in $E$ is used to recognize the stability and the concrete usage is described in detail subsequently.
}%
{%
  本研究の能動振動を用いた安定性認識の手順は以下のようになっている(\figref{figure:active-vibration}).
  \begin{enumerate}
    \item 振動させる筋のグループ$V$, 振動を観測する筋のグループ$O$を決める.
    \item $V$の筋の筋長制御入力$\bm{l}^{ref}$を振動させる.
    \item $O$の筋の筋張力$\bm{f}$から振動の伝播度合いを検出する.
  \end{enumerate}

  1)について, $V$と$O$の選び方はロボットによって異なるため, \secref{sec:preliminary-experiments}にてその選び方については考察を行う.
  また, $M_{\{V, O\}}$を$V$または$O$に含まれる筋の数, $\bm{l}^{ref}_{\{V, O\}}$を$V$または$O$に含まれる筋の指令筋長, $\bm{f}_{\{V, O\}}$を$V$または$O$に含まれる筋の筋張力とする.

  2)について, $V$に含まれるそれぞれの筋の指令筋長$l^{ref}_{V}$を以下に示すように変化させる.
  \begin{align}
    l^{ref'}_{V} = l^{ref}_{V} + A\textrm{sin}(2{\pi}Ft)
  \end{align}
  ここで, $t$は現在時刻, $F$は振動の周波数, $A$は振動の振幅, $\bm{l}^{ref}_{V}$は元々の指令筋長, $\bm{l}^{ref'}_{V}$は振動を追加して実機に送られる指令筋長を表す.

  3)について, $O$に含まれる筋の筋張力$\bm{f}_{O}$を用いて, 以下のように評価値$E$を計算する.
  \begin{align}
    \bm{h} &= \textrm{Extract}_{[F_{min}, F_{max}]}(\textrm{FFT}(\bm{f}_{O, [t-T+1, t]})) \label{eq:fft}\\
    E_{raw} &= \frac{1}{M_{O'}}\sum_{i \in O'} h_{i} \label{eq:ave}\\
    E &= (1-{\alpha})E+{\alpha}E_{raw} \label{eq:lowpass}
  \end{align}
  ここで, $\textrm{FFT}$はフーリエ変換, $T$はフーリエ変換を行うデータの長さ, $\bm{f}_{O, [t-T+1, t]}$は$[t-T+1, t]$の範囲の$\bm{f}_{O}$を並べた時系列データ, $\textrm{Extract}_{[F_{min}, F_{max}]}$は指定された周波数領域$[F_{min}, F_{max}]$で最大のフーリエ変換後の値を取り出す関数を表す.
  つまり, \equref{eq:fft}は, $O$に含まれるそれぞれの筋について, 個別にフーリエ変換し, 決められた周波数帯の中で最も大きなスペクトルを取り出す操作である.
  個別にフーリエ変換せずに$\bm{f}_{O}$のL2ノルムのみをフーリエ変換することも考えられるが, 振動は同位相でもそれぞれの観測される筋で位相が異なることがあったため, このような形としている.
  また, $O'$は, $O$から$h$の値が大きい筋を上から2つ, 小さい筋を上から2つ抜いたグルーピングである.
  つまり, \equref{eq:ave}は, $O'$に含まれる筋の$h$の平均を表す.
  この\equref{eq:ave}は実験を通して見つけたヒューリスティックであり, 環境との接触によって外力が加わった際, 一部の筋に負荷がかかり, その負荷が振動の伝播度合いを変化させてしまうことがあったためである.
  この操作を施すことで, 環境からの負荷ではなく, 純粋に環境による振動の抑制を扱うことが可能である.
  筋数が少ない場合はこの操作を抜き, $O$に含まれる全ての筋について$h$の平均を取ることも考えられ, 性能は落ちるが大きく特性が変わることはない.
  最後に, $\alpha$を係数として, \equref{eq:lowpass}でlow pass filterをかける.
  この$E$の変化を用いて安定性を認識するが, 具体的な使い方については後に詳細を述べる.
}%

\section{Preliminary Experiments of Active Vibration and Stability Recognition} \label{sec:preliminary-experiments}
\switchlanguage%
{%
  In the experiments of this section, we compare the differences of $E$ when changing the environment (Env), amplitude (Amp), frequency (Freq), location of vibration $V$ and observation $O$ (Group), and the arm posture (Posture), and we find the way of using $E$ to evaluate the stability of the robot.
  The fixed and changed parameters for all experiments is shown in \tabref{table:preliminary-experiments}.
}%
{%
  本節の実験は, 環境(Env)による違い, 振幅(Amp)による違い, 周波数(Freq)による違い, 振動箇所$V$と振動を観測する箇所$O$ (Group)による違い, 腕の姿勢(Posture)による違い, について比較実験を行い, 安定性評価への利用法を見出す.
  \tabref{table:preliminary-experiments}に全実験の固定パラメータと変化パラメータに関する表を示す.
}%

  \begin{table}[htb]
    \centering
    \caption{Preliminary experimental configurations. ``C'' means changing the parameter, ``+'' means changing the parameter partially, and ``-'' means fixing the parameter.}
    \begin{tabular}{l|ccccc}
      & Env & Amp & Freq & Group & Posture\\ \hline
      Exp-1 & C & - & - & - & + (local) \\
      Exp-2 & + & C & - & - & + (local) \\
      Exp-3 & + & - & C & - & + (local) \\
      Exp-4 & + & - & - & C & + (local) \\
      Exp-5 & + & - & - & - & C (global) \\
    \end{tabular}
    \label{table:preliminary-experiments}
  \end{table}

\begin{figure}[t]
  \centering
  \includegraphics[width=0.7\columnwidth]{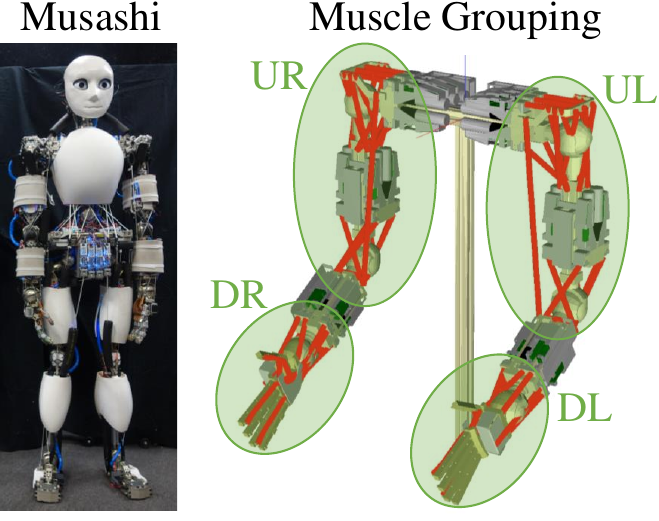}
  \vspace{-1.0ex}
  \caption{The musculoskeletal humanoid Musashi and its muscle grouping used in this study.}
  \label{figure:musashi-setup}
  \vspace{-3.0ex}
\end{figure}

\begin{figure}[t]
  \centering
  \includegraphics[width=0.8\columnwidth]{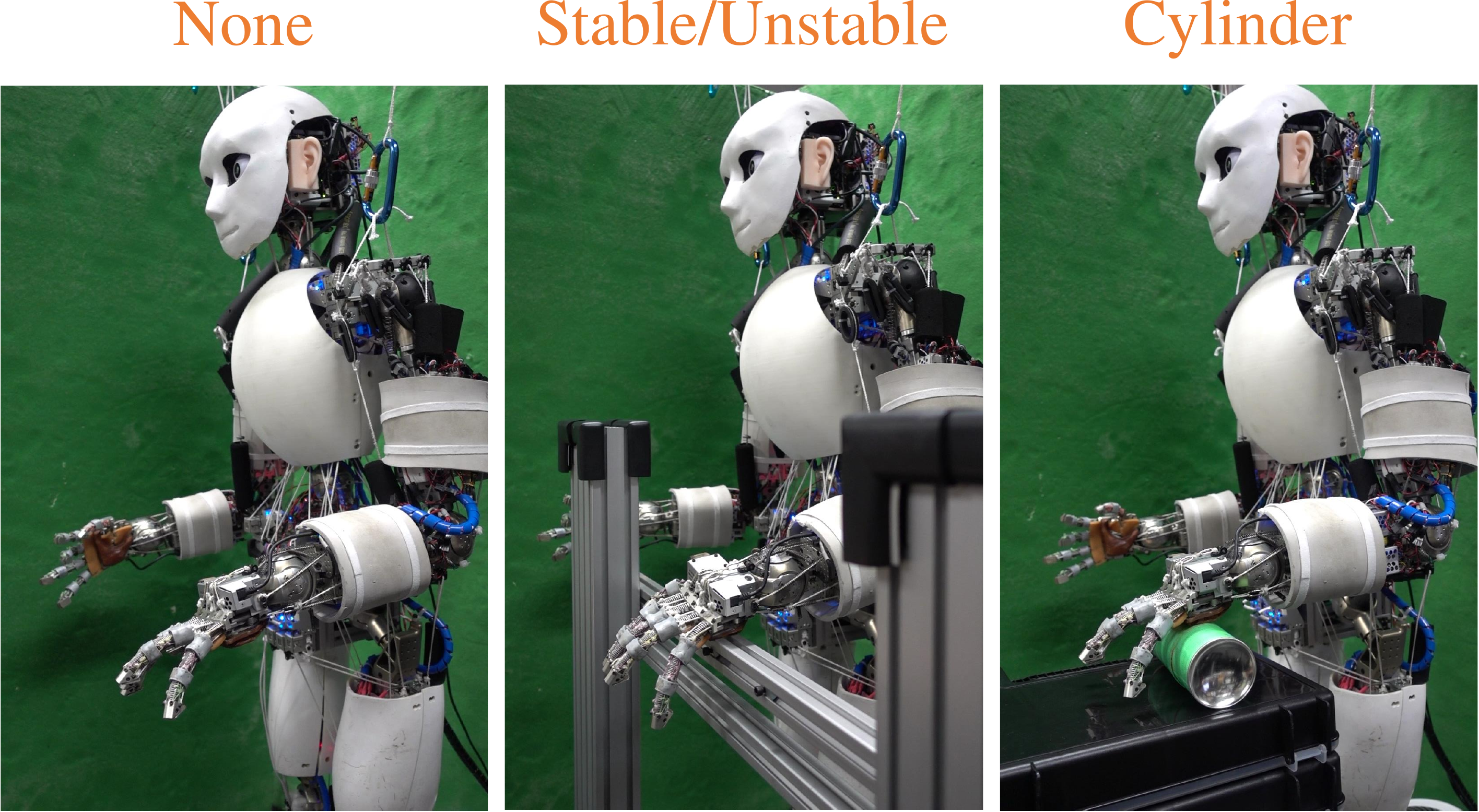}
  \vspace{-1.0ex}
  \caption{Experimental setup of environments used in this study.}
  \label{figure:environmental-setup}
  \vspace{-3.0ex}
\end{figure}

\subsection{Experimental Setup} \label{subsec:experimental-setup}
\switchlanguage%
{%
  In this study, we use the musculoskeletal humanoid Musashi \cite{kawaharazuka2019musashi} (\figref{figure:musashi-setup}).
  Since there are innumerable ways to choose $V$ and $O$ muscle groups, we divided all the muscles of both arms into four groups in this study.
  The left upper arm is named $UL$, the left forearm $DL$, the right upper arm $UR$, and the right forearm $DR$, respectively, as shown in \figref{figure:musashi-setup}.
  This is an independent grouping in which there are no articulated muscles between the groups.
  Each upper arm group contains ten muscles including biarticular muscles and each forearm group contains eight muscles.
  If such a grouping is difficult due to the large number of articulated muscles, a grouping method such as \cite{kawaharazuka2018estimator} may be useful.

  The control cycle of the robot is 125 Hz, and the process of \equref{eq:fft}--\equref{eq:lowpass} is executed at 20 Hz.
  Also, we set $T=40$, $F_{min}=F-2$, $F_{max}=F+2$, and $\alpha=0.1$.

  In this experiment, the robot moves mainly its left elbow in the standing position and touches the environment with its left hand.
  The joint angle command is performed feedforwardly by means of \cite{kawaharazuka2019longtime}, but due to its flexible body, the command value is not always achievable when it contacts the environment.
  In this case, four types of environments are prepared as shown in \figref{figure:environmental-setup}.
  They are: no environment (None), a rigid frame (Stable) as shown in the middle figure, the unscrewed rigid frame (Unstable), and a cylindrical rigid object placed on a table as shown in the right figure.
  The height of the environment is adjusted to be the same in each case.
  These four types are compared with each other in \secref{subsec:environment-exp}, and the other experiments are conducted with respect to None and Stable.
}%
{%
  本研究では筋骨格ヒューマノイドMusashi \cite{kawaharazuka2019musashi}を用いる(\figref{figure:musashi-setup}).
  $V$と$O$の筋群の選び方は無数にあるため, 本研究では両腕に4つのグルーピングを作成した.
  それぞれ図に示すように, 左上腕を$UL$, 左前腕を$DL$, 右上腕を$UR$, 右前腕を$DR$と名づけている.
  これは, そのグルーピング間を多関節筋が跨がないような独立したグループである.
  上腕のグループには2関節筋を含むそれぞれ10本ずつの筋が, 前腕のグループにはそれぞれ8本ずつの筋が含まれている.
  多関節筋が多くこのようなグルーピングが難しい場合は, \cite{kawaharazuka2018estimator}のようなグルーピングも考えられる.

  ロボットの動作周期は125 Hzであり, \equref{eq:fft}--\equref{eq:lowpass}の処理は20 Hzで実行される.
  また, $T=40$, $F_{min}=F-2$, $F_{max}=F+2$, $\alpha=0.1$としている.

  本実験では, ロボットが直立した状態で主に左肘を動かし, 左手を環境に接触させる.
  関節角度指令は\cite{kawaharazuka2019longtime}によってフィードフォワードに行うが, 柔軟な身体ゆえに, 特に環境と接触する際は指令値を実現できるわけではない.
  このとき, 環境は4種類用意した(\figref{figure:environmental-setup}).
  それらは, 環境なし(None), 中図にあるような剛体フレーム(Stable), この剛体フレームのネジを緩めたもの(Unstable), 円筒の剛体を机の上に図のように置いたものである.
  それぞれ環境の高さは同じになるように調整されている.
  \secref{subsec:environment-exp}においてはこれら4種類の比較について行い, その他の実験についてはNoneとStableに関して実験を行う.
}%

\begin{figure*}[t]
  \centering
  \includegraphics[width=2.0\columnwidth]{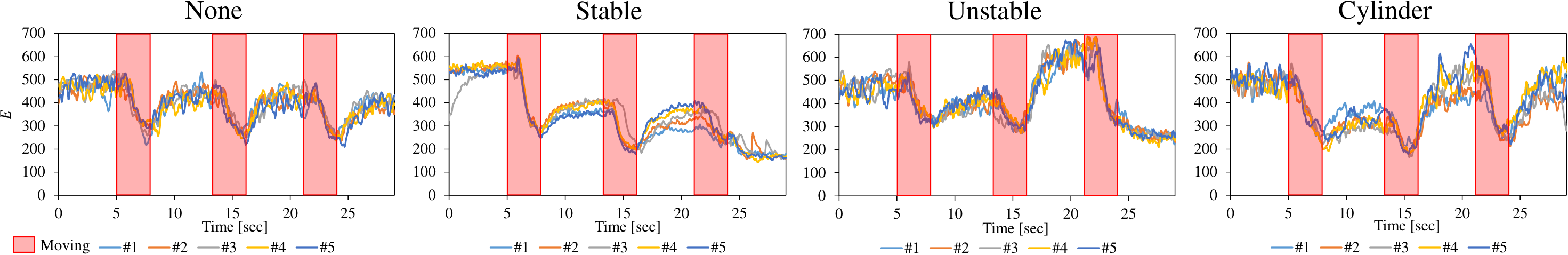}
  \vspace{-1.0ex}
  \caption{Exp-1: Comparison of 5 transitions of $E$ among four environments: None, Stable, Unstable, Cylinder. The red shading indicates that the robot is moving.}
  \label{figure:environment-graph1}
  \vspace{-1.0ex}
\end{figure*}

\begin{figure}[t]
  \centering
  \includegraphics[width=0.8\columnwidth]{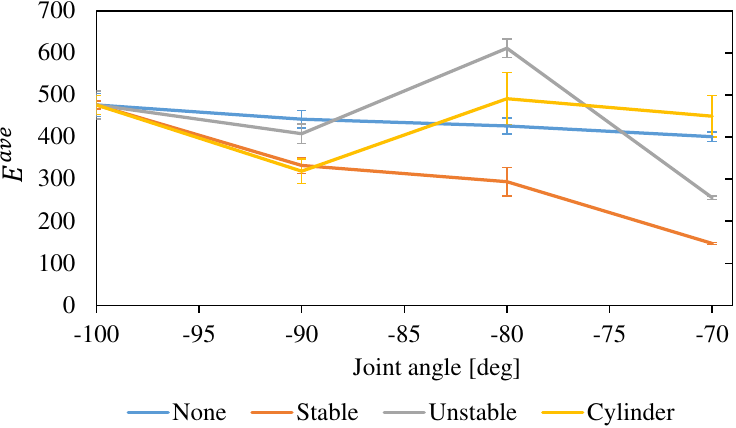}
  \vspace{-1.0ex}
  \caption{Exp-1: Comparison of average and standard deviation of 5 $E^{ave}$ transitions among four environments.}
  \label{figure:environment-graph2}
  \vspace{-1.0ex}
\end{figure}

\begin{figure}[t]
  \centering
  \includegraphics[width=1.0\columnwidth]{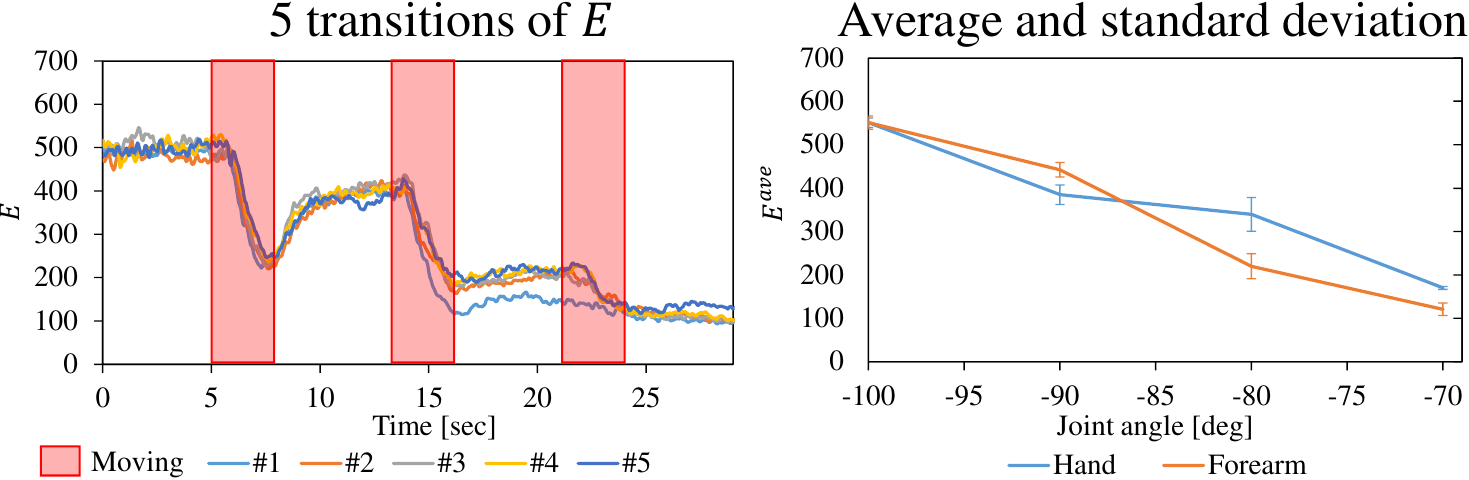}
  \vspace{-3.0ex}
  \caption{Exp-1: The left graph shows 5 transitions of $E$ when contacting Stable by the forearm. The right graph shows the comparison of $E^{ave}$ when contacting Stable by the hand or the forearm.}
  \label{figure:environment-graph3}
  \vspace{-3.0ex}
\end{figure}

\subsection{Exp-1: Changing Environment} \label{subsec:environment-exp}
\switchlanguage%
{%
  In this experiment, we fix the grouping, frequency, and amplitude, and locally move the posture to observe the changes in $E$ when contacting four environments of \secref{subsec:experimental-setup}.
  $V$ is set to $UR$, $O$ is set to $DL$, and we set $F=12.5$ and $A=3$.
  The right elbow is bent at -90 deg and the left elbow is lowered from -100 to -70 deg.
  From -100 to -90, from -90 to -80, and from -80 to -70 deg, the robot moves over 3 seconds and then it stops for 5 seconds, repeatedly.
  The change of $E$ during the same 5 movements for each environment is shown in \figref{figure:environment-graph1}.
  We also take the mean $E^{ave}$ of $E$ for the last second after the left elbow angle is at -100, -90, -80 and -70 deg for 5 seconds and show the mean and variance of $E^{ave}$ during the 5 movements in \figref{figure:environment-graph2}.
  As can be seen from \figref{figure:environment-graph1}, there is a certain degree of variance but the values are reproducible for the same movement.
  From \figref{figure:environment-graph2}, we can see that $E^{ave}$ does not change significantly in the case of None, $E^{ave}$ gradually decreases in the case of Stable, and $E^{ave}$ gradually decreases after increasing once in the case of Unstable.
  In the case of Cylinder, $E^{ave}$ goes down initially but does not go down further when the hand is pressed to the environment, showing a value equal to or higher than None.
  Overall, $E^{ave}$ tends to decrease in stable environments, and tends to be the same as or larger than the value without contact to the environment in unstable environments.
  In the case of Unstable, we can assume that the body becomes unstable (high $E^{ave}$) when the hand is not pressed too hard, but shifts to the stable state (low $E^{ave}$) when the hand is pressed firmly.

  Also, although the hand contacts the environment for the above experiments, we conducted the same experiment with the forearm in contact regarding Stable.
  As shown in \figref{figure:environment-graph3}, we can confirm that similar behavior to the case where the hand contacts the environment can be seen.
}%
{%
  本実験では, グルーピング・周波数・振幅を固定し, 姿勢を局所的に徐々に動かして\secref{subsec:experimental-setup}の4つの環境に接触する際の$E$の変化を観察する.
  $V$を$UR$, $O$を$DL$に固定し, $F=12.5$, $A=3$とした.
  姿勢は, 右肘を-90度に曲げた状態で, 左肘を-100度から-70度まで下ろしていく.
  -100から-90, -90から-80, -80から-70へと, 3秒で動作し, その後5秒間止まる, という動作を繰り返す.
  それぞれの環境について同じ動作を5回行った際の$E$の変化を\figref{figure:environment-graph1}に示す.
  また, 左肘の角度が-100, -90, -80, -70度で5秒間止まった際の最後の1秒について$E$の平均$E^{ave}$を取り, $E^{ave}$の5回動作中値の平均と分散を\figref{figure:environment-graph2}に示す.
  \figref{figure:environment-graph1}からわかるように, ある程度の分散はあるものの, 同じ動作をする分には値に再現性があることがわかる.
  \figref{figure:environment-graph2}からは, 環境に身体を押し付けるに従って, Noneの場合は大きく$E^{ave}$は変化しない, Stableの場合は徐々に$E^{ave}$が下がっていく, Unstableの場合は一度$E^{ave}$は上がるものの, その後下がっていく, Cylinderの場合は最初は$E^{ave}$が下がったものの, 押し付けてもそれ以上下がることはなく, Noneと同程度からそれ以上の値を示している.
  全体的な傾向として, 安定した環境では$E^{ave}$は下がる傾向にあり, 不安定な環境では$E^{ave}$は環境と接触していない状態と同じかそれよりも大きな値となる傾向にある.
  Unstableについては, 強く押し付けない状態では不安定になるものの, 強く押し付けることによって安定した状態へと移行したと考えられる.

  また, 上記の実験は手が環境に接触しているが, Stableについて, 前腕を接触させたときにも同様に実験を行った結果を\figref{figure:environment-graph3}に示す.
  右図からわかるように, 手を接触させた場合と似た挙動を確認することができた.
}%

\begin{figure}[t]
  \centering
  \includegraphics[width=1.0\columnwidth]{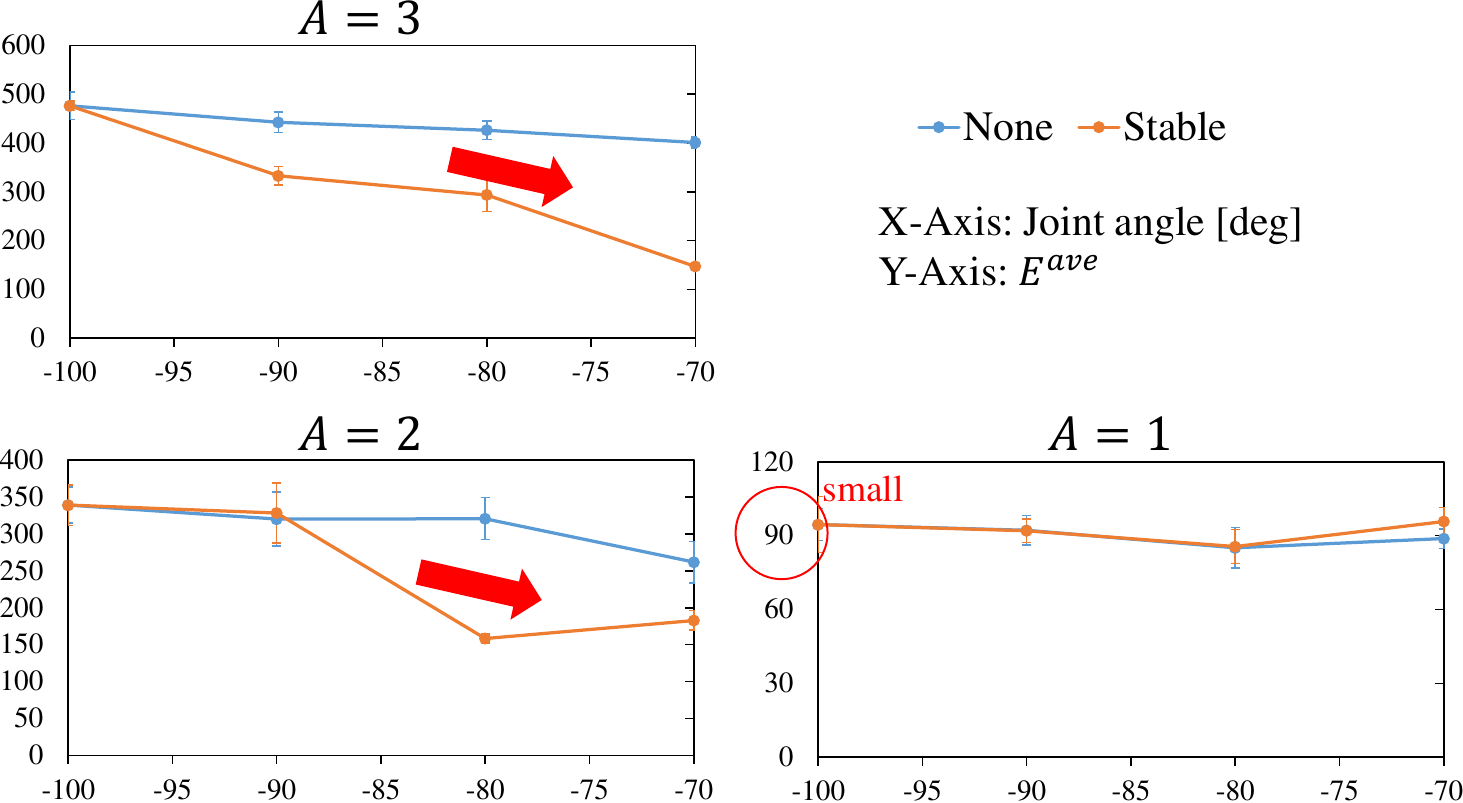}
  \vspace{-3.0ex}
  \caption{Exp-2: Comparison of $E^{ave}$ when contacting None/Stable and changing $A$.}
  \label{figure:amplitude-graph1}
  \vspace{-1.0ex}
\end{figure}

\subsection{Exp-2: Changing Amplitude} \label{subsec:amplitude-exp}
\switchlanguage%
{%
  In this experiment, we fix the grouping and frequency, and observe the change of $E$ when contacting None/Stable environment by locally changing the posture, while changing the amplitude.
  $V$ is set to $UR$, $O$ is set to $DL$, and we set $F=12.5$.
  The amplitudes $A$ of 1, 2, and 3 [mm] are compared.
  The change of $E^{ave}$ when conducting the same movement as in \secref{subsec:environment-exp} is shown in \figref{figure:amplitude-graph1}.

  At $A=2$ and $A=3$, $E^{ave}$ is lowered by pressing the hand to Stable compared to None, whereas at $A=1$, there is almost no difference between None and Stable.
  As $A$ is decreased, the value of $E^{ave}$ decreases to about 500, 300, and 100.
  When $A=0$, the average value of $E^{ave}$ in 10 seconds is 71.1, which means that the state of $A=1$ is not much different from the state of $A=0$ without vibration.
  Also, it is difficult to make $A$ larger than $A=3$ because the robot vibrates more intensely as $A$ is increased.
}%
{%
  本実験では, グルーピング・周波数を固定し, 姿勢を局所的に徐々に動かしてNone/Stableの環境に接触する際の$E$の変化を, 振幅を変化させながら観察する.
  $V$を$UR$, $O$を$DL$に固定し, $F=12.5$とした.
  振幅$A$は, 1, 2, 3 [mm]でそれぞれ比較する.
  \secref{subsec:environment-exp}と同様の動作を行った際の$E^{ave}$の変化を\figref{figure:amplitude-graph1}に示す.
  $A=2, 3$のときは両者ともStableに押し付けることでNoneに比べて大きく$E^{ave}$が下がっており, これに対して$A=1$ではNoneとStableにほとんど差がないことがわかる.
  また, $A$を下げるごとに$E^{ave}$の値は約500, 300, 100程度に全体的に下がっていく.
  $A=0$, つまり振動させていないときにおける$E^{ave}$の10秒間の平均を取ったところ値は71.1であり, $A=1$の状態は振動させていない状態と大きな違いがない.
  これに対して, $A$を大きくするに連れて徐々にロボットの振動が激しくなるため, $A=3$より大きくすることは難しい.
}%

\begin{figure}[t]
  \centering
  \includegraphics[width=1.0\columnwidth]{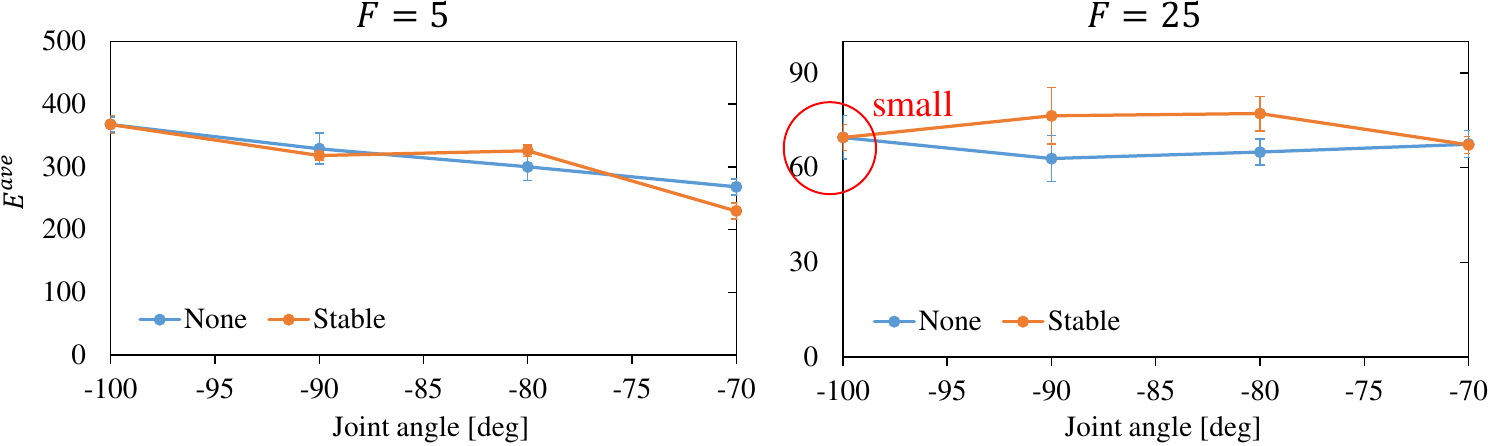}
  \vspace{-3.0ex}
  \caption{Exp-3: Comparison of $E^{ave}$ when contacting None/Stable and changing $F$.}
  \label{figure:frequency-graph1}
  \vspace{-3.0ex}
\end{figure}

\begin{figure*}[t]
  \centering
  \includegraphics[width=2.0\columnwidth]{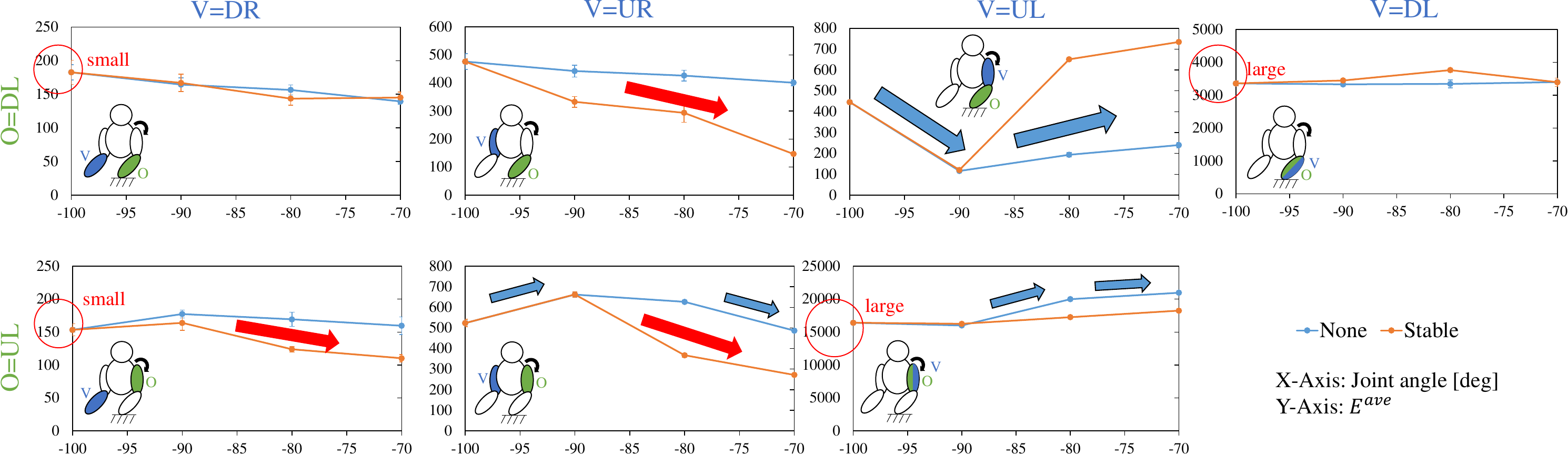}
  \vspace{-1.0ex}
  \caption{Exp-4: Comparison of $E^{ave}$ when contacting None/Stable and changing Grouping of $V$ and $O$.}
  \label{figure:grouping-graph1}
  \vspace{-1.0ex}
\end{figure*}

\subsection{Exp-3: Changing Frequency} \label{subsec:frequency-exp}
\switchlanguage%
{%
  In this experiment, we fix the amplitude and grouping, and observe the change of $E$ when contacting None/Stable environment by locally changing the posture, while changing the frequency.
  $V$ is set to $UR$, $O$ is set to $DL$, and we set $A=3$.
  The frequency $F$ is compared with 5 and 25 [Hz] (for $F=12.5$, see the upper left figure $A=3$ of \figref{figure:amplitude-graph1}).
  The change of $E^{ave}$ when conducting the same movement as in \secref{subsec:environment-exp} is shown in \figref{figure:frequency-graph1}.
  For $F=5$, there is no significant change in $E^{ave}$.
  This is considered to be due to the fact that it is difficult to calculate the characteristics at a small frequency $F$ with respect to the data length $T$ of FFT.
  To solve this problem, we can make $T$ longer, but there are some problems such as the calculation time and the long time required for the true change of $E$ to appear after contact with the environment.
  For $F=25$, the value of $E^{ave}$ is quite small, about 70, and is indistinguishable from noise, as is $A=1$ in \secref{subsec:amplitude-exp}.
  This is probably due to the fact that when $F$ is large, the length of the muscles cannot follow the length of $\bm{l}^{ref'}$ and the vibrations become small.
}%
{%
  本実験では, 振幅・グルーピングを固定し, 姿勢を局所的に徐々に動かしてNone/Stableの環境に接触する際の$E$の変化を, 周波数を変化させながら観察する.
  $V$を$UR$, $O$を$DL$に固定し, $A=3$とした.
  周波数$F$は, 5, 25 [Hz]で実験をする($F=12.5$の場合は\figref{figure:amplitude-graph1}の左上$A=3$を参照).
  \secref{subsec:environment-exp}と同様の動作を行った際の$E^{ave}$の変化を\figref{figure:frequency-graph1}に示す.
  $F=5$のときは, $E^{ave}$に大きな変化は見られなかった.
  これは, FFTを行う時系列の長さ$T$に対して小さな周波数$F$における特性を計算するのが難しくなるからであると考えられる.
  これを解決するには$T$を長く取る等が考えられるが, 計算に時間がかかる, 環境と接触してから真の$E$の変化が現れるまでに時間を要する, 等の問題がある.
  また, $F=25$のとき, $E^{ave}$の値は70程度とかなり小さく, \secref{subsec:amplitude-exp}の$A=1$と同様, ノイズと区別がつかない.
  これは, $F$が大きいと$\bm{l}^{ref'}$に筋の長さが追従できなくなり, 振動が小さくなり変化がわかりにくくなってしまっていることが原因だと考えられる.
}%

\subsection{Exp-4: Changing Grouping} \label{subsec:grouping-exp}
\switchlanguage%
{%
  In this experiment, we fix the frequency and amplitude, and observe the change of $E$ when contacting None/Stable environment by locally changing the posture, while changing the grouping.
  We set $F=12.5$ and $A$ is determined manually according to the grouping specification.
  As the left hand is assumed to touch the environment, $O$ is restricted to $\{UL, DL\}$.
  When $O$ is $DL$ and $V$ is $\{DR, UR, UL, DL\}$, and when $O$ is $UL$ and $V$ is $\{DR, UR, UL\}$, the change of $E$ when conducting the same movement as in \secref{subsec:environment-exp} is shown in \figref{figure:grouping-graph1}.
  Since the closer $O$ and $V$ are, the more easily vibrations are propagated, $A$ is set to 3, 3, 2, and 1 for $V$ when $V$ is $\{DR, UR, UL, DL\}$, respectively.

  As an overall trend, there is no significant difference between None and Stable for $(O=DL, V=DR)$, $(O=V=DL)$, and $(O=V=UL)$, and except for $(O=DL, V=UL)$, $E^{ave}$ is smaller for Stable than for None.
  From the results, first, when $V$ and $O$ are identical, such as $(O=V=DL)$ and $(O=V=UL)$, $E^{ave}$ is extremely large and there is no significant difference between None and Stable.
  This is because $E^{ave}$ becomes large due to direct transmission of vibrations, and at the same time, it is considered that $E^{ave}$ does not respond well to the change in the environment.
  Second, regarding $(O=DL, V=UL)$, the value of $E^{ave}$ changes significantly even for None without any contact.
  The reason for this is that $UL$ is the part that is locally moved in the experiment, and the friction state changes significantly during the evaluation when $V=UL$.
  For $(O=V=UL)$, there is a slight change in $E^{ave}$ at None, but the change is considered to be smaller than for $(O=DL, V=UL)$, since the vibration is directly observed and is not easily affected by friction.
  In addition, $E^{ave}$ is more likely to change at None for $O=UL$ than for $O=DL$.
  Third, when $V=DR$, the value of $E^{ave}$ is quite small, around 150.
  This is considered to be due to the fact that $V$ and $O$ are distant from each other and the vibrations are not well propagated.
  Finally, regarding $(O=DL, V=UR)$, $(O=UL, V=UR)$, and $(O=UL, V=DR)$, $E^{ave}$ with Stable is lower than that with None, and it can be used to recognize the stability.
}%
{%
  本実験では, 周波数・振幅を固定した状態で, 姿勢を局所的に徐々に動かしてNone/Stableの環境に接触する際の$E$の変化を観察する.
  $F=12.5$, $A$はグルーピングの指定に応じて手動で決定している.
  左手が環境に触れる前提のため, $O$は$UL$, または$DL$に絞る.
  $O$を$DL$として$V$を$DR, UR, UL, DL$としたとき, $O$を$UL$として$V$を$DR, UR, UL$としたとき, \secref{subsec:environment-exp}と同様の動作を行った際の$E$の変化を\figref{figure:grouping-graph1}に示す.
  $O$と$V$は近いほど振動が伝わりやすいため, $V$が$DR, UR, UL, DL$のとき, $A$はそれぞれ$3, 3, 2, 1$を用いる.

  全体的な傾向として, $(V=DR, O=DL)$, $(V=O=DL)$, $(V=O=UL)$のときはNone/Stableで大きな違いが見られず, その他では$O=DL, V=UL$を除いて, Stableの方がNoneに比べて$E^{ave}$が小さくなっていることがわかる.
  まず, $(V=O=DL)$や$(V=O=UL)$のように, $V$と$O$が一致する場合には, $E^{ave}$が極端に大きな値となり, None/Stableにおいて大きな違いがないことがわかる.
  これは, 振動が直接伝わってしまうため$E^{ave}$が大きくなり, 同時に環境という外部からの接触による違いに反応しにくくなってしまっていると考えられる.
  次に, $(O=DL, V=UL)$のときは, 環境に触れないNoneでさえ$E^{ave}$の値が大きく変化してしまっている.
  これは, 実験で局所的に動かす部位が$UL$であるため, $V=UL$としてしまうと, 評価中に摩擦等の状態, そして$E$が大きく変化してしまうことが原因だと考えられる.
  $(O=V=UL)$についても同様にNoneにおいて$E^{ave}$の変化が多少見られるが, 振動を直接観測し摩擦等の影響を受けにくいため, $(O=DL, V=UL)$のときよりも変化が小さいと考えられる.
  この他, $O=UL$のグループについてもそれぞれ, $O=DL$のグループに比べると, Noneにおいて$E^{ave}$が変化しやすくなっている.
  次に, $V=DR$のとき, $E^{ave}$の値は150程度とかなり小さい.
  これは, $V$と$O$が離れており, 振動が上手く伝達しなかったためであると考えられる.
  最後に, $(V=UR, O=DL)$, $(V=UR, O=UL)$, $(V=DR, O=UL)$では環境によりNoneに比べてStableの$E^{ave}$がしっかりと下がっており, 安定性を検知することが可能であると考えられる.
}%

\begin{figure}[t]
  \centering
  \includegraphics[width=1.0\columnwidth]{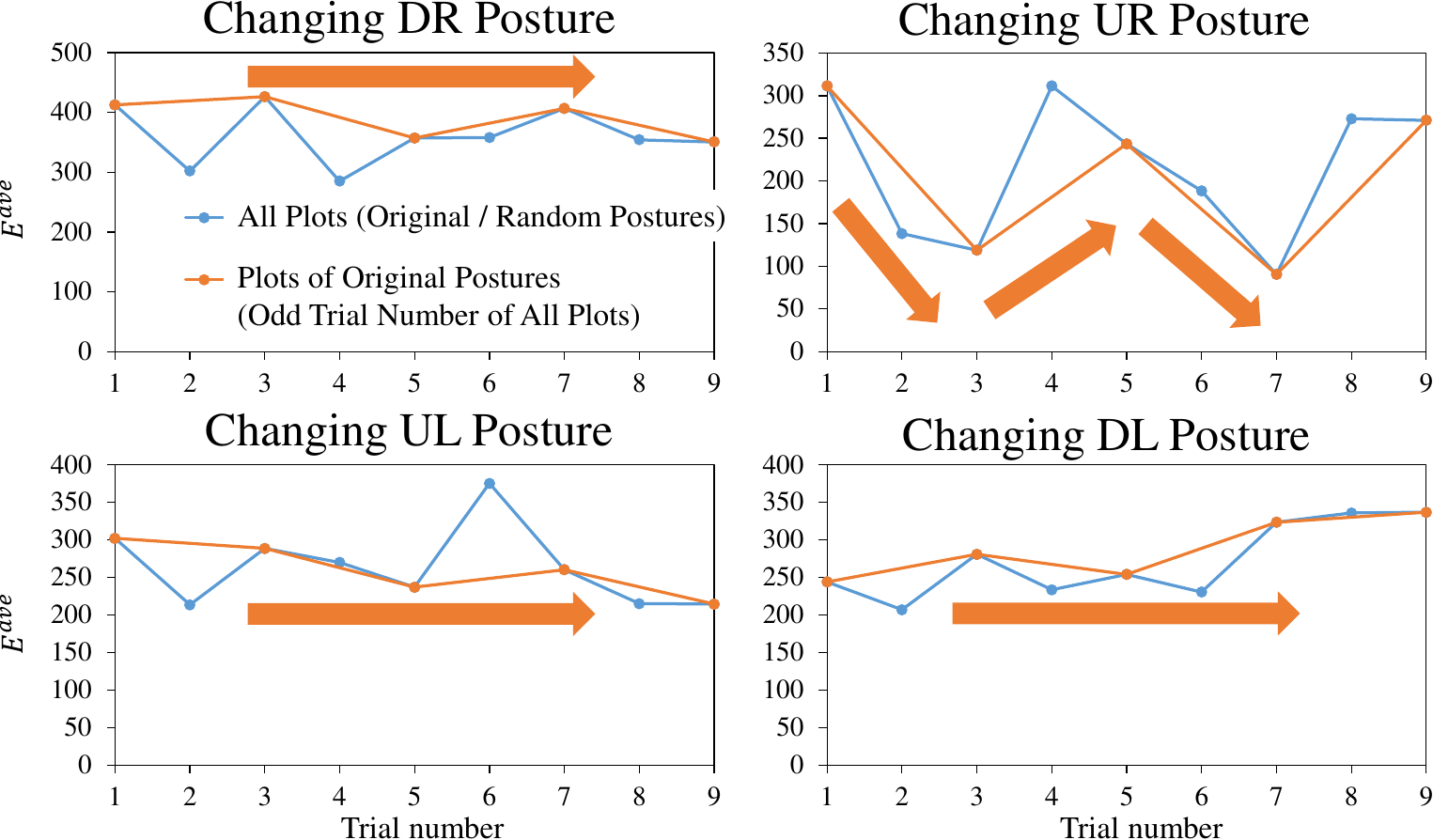}
  \vspace{-3.0ex}
  \caption{Exp-5: Comparison of $E^{ave}$ when moving $DR$, $UR$, $UL$, or $DL$ to the initial pose (odd trial numbers) and random joint angles (even trial numbers) alternately without contacting any environment.}
  \label{figure:posture-graph1}
  \vspace{-3.0ex}
\end{figure}

\subsection{Exp-5:  Changing Posture} \label{subsec:posture-exp}
\switchlanguage%
{%
  In this experiment, we fix the grouping, amplitude, and frequency, and observe the change of $E$ when globally changing the posture.
  The setting is basically the same as that of \secref{subsec:environment-exp}, but we observe the behavior of $E$ when each $\{DR, UR, UL, DL\}$ is moved globally within the range of joint angle instead of locally.
  \figref{figure:posture-graph1} shows the change of $E$ in None when each $\{DR, UR, UL, DL\}$ is moved randomly within the range of motion and returned to the initial posture (the joint angles of the left and right elbows are -90 deg) repeatedly.
  When the trial number is odd, it represents the initial posture, and when it is even, it represents a random posture.
  In any case, the global movement changes $E^{ave}$ even when the robot is not in contact with the environment, which means that there is a postural dependency.
  Also, when $UR$, i.e. $V$, is moved, the value of $E^{ave}$ does not remain the same even after returning to the same initial posture, but returns to roughly the same value when other groups are moved.
  This may be due to the fact that when $V$ is moved, the effects of friction and other factors are changed and hysteresis is generated in the value of $E^{ave}$.
}%
{%
  本実験では, 周波数・振幅・グルーピングを固定した状態で, 姿勢を大域的に動かした際の$E$の変化を観察する.
  設定は\secref{subsec:environment-exp}と基本的には同じであるが, 局所的ではなく, 大域的にそれぞれ$DR, UR, UL, DL$を動作させたときの$E$の挙動を観察する.
  $DR, UR, UL, DL$の姿勢を可動域の中でランダムに動かし, 初期姿勢に戻す(右肘と左肘の角度が-90度の状態)ことを繰り返したときのNoneにおける$E$の変化を\figref{figure:posture-graph1}に示す.
  Trialの番号が奇数の時は初期姿勢, 偶数の時はランダムな姿勢を表す.
  どの場合に置いても, 大域的に動かした場合, 環境に接触していなくても$E^{ave}$が変化する, つまり, 姿勢依存性があることがわかる.
  また, $UR$, つまり$V$を動かした場合は, 同じ初期姿勢に戻っても$E^{ave}$の値は同じにならず, その他のグループを動かした場合は大体同じ値に戻ってくることがわかる.
  これは, $V$を動作させてしまうと摩擦等の影響が変化し, $E^{ave}$の値にヒステリシスが生まれてしまうことが原因であると考えられる.
}%


\subsection{Summary} \label{subsec:summary}
\switchlanguage%
{%
  Our findings from these experiments can be summarized as follows.
  \begin{itemize}
    \item The value of $E$ tends to decrease when contacting the stable environment and to increase when contacting the unstable environment.
    \item If the amplitude $A$ is too small, $E$ does not change when contacting any environment, and if $A$ is too large, the robot will vibrate too strongly, so it should be set appropriately.
    \item If the frequency $F$ is too small, it is difficult to detect the change in $E$ because it is mixed with noise, and if $F$ is too large, it is impossible to follow the commanded muscle length and the vibration is lost.
    \item If $O$ and $V$ are too distant to each other, the vibration will not be transmitted well, and if they are too close to each other, the vibration and observation will be directly connected to each other, so they should be appropriately arranged.
    \item If muscles included in $O$ and $V$ are moved, $E$ may change even if it is not in contact with the environment, so it should be avoided if possible.
    \item If the posture is globally changed, $E$ will change without contact with the environment, and especially, global movement of the muscles belonging to $V$ should be avoided because the effects of friction changes and hysteresis is created in $E$.

  \end{itemize}

  In this study, we set $V=UR$, $O=DL$, $A=3$, and $F=12.5$, which is the same as \secref{subsec:environment-exp}, for subsequent experiments.
  The value of $E^{ave}$ cannot be deduced from the robot posture because hysteresis is created in $E^{ave}$ even in the same posture depending on the moved part of the body.
  Also, the value of $E^{ave}$ will be changed for None if the muscles in $V$ and $O$ are moved or if the global posture is changed.
  Therefore, we can recognize the stability of $E^{ave}$ by locally moving the region that does not belong to $O$ or $V$ and comparing the value of $E^{ave}$ before and after moving the region.
  The following $S$ is used as the final evaluation value,
  \begin{align}
    S = E^{ave}_{2} / E^{ave}_{1}
  \end{align}
  where $E^{ave}_{\{1, 2\}}$ refers to $E^{ave}$ before and after a local posture change, respectively.
}%
{%
  これらの実験からわかったことは以下のようにまとめられる.
  \begin{itemize}
    \item 評価値$E$は安定した環境に接触すると下がり, 不安定な環境に接触すると上がる傾向にある.
    \item 振幅$A$は小さすぎると環境に接触した際の$E$に変化が出なくなり, 大きすぎるとロボットが強く振動してしまうため, 適切に設定する必要がある.
    \item 周波数$F$は小さすぎるとノイズと混ざって変化を検知しにくくなり, 大きすぎると指令筋長に追従できないため振動がなくなり$E$の変化が検知できなくなるため, 適切に設定する必要がある.
    \item グルーピング$O$と$V$は離れすぎると振動が伝わらなくなり, 近すぎると振動と観測が直結し環境接触による違いが反映されにくくなってしまうため, 適切に離して配置する必要がある.
    \item グルーピング$O$や$V$に選ばれた筋を動かすと環境に接触していなくても$E$が変化してしまうことがあるため, なるべく避けるべきである.
    \item 姿勢を大局的に動かすと環境に接触しなくても$E$が変化し, 特に$V$に属する筋を大局的に動かすと摩擦等の影響が変化し$E$にヒステリシスが生まれるため避けるべきである.
  \end{itemize}

  これらから, 本研究では$V=UR, O=DL, A=3, F=12.5$のように設定する(\secref{subsec:environment-exp}と同じ)に設定する.
  動作させる部位によっては同じ姿勢においても$E^{ave}$にヒステリシスが生まれてしまうため, 姿勢から$E^{ave}$の値を推論することはできない.
  また, $V$や$O$のグルーピングを動かしてしまう, または大域的姿勢を変化させるとNoneにおいても$E^{ave}$の値は変化してしまう.
  そこで, $O$や$V$に属さない部位を局所的に動かし, 動かす前と後の$E^{ave}$を比べることで, 安定性を認識する.
  つまり, 以下の値$S$を最終的な評価値として用いる.
  \begin{align}
    S = E^{ave}_{2} / E^{ave}_{1}
  \end{align}
  ここで, $E^{ave}_{\{1, 2\}}$はそれぞれ局所的に姿勢を変化させる前と後における$E^{ave}$を指す.
}%

\section{Motion Extension and Bracing Behavior Using Environment} \label{sec:experiments}
\switchlanguage%
{%
  Based on the results obtained in \secref{subsec:summary}, we conducted experiments for the motion extension and bracing behavior.
}%
{%
  \secref{subsec:summary}で得られた結果をもとに, 動作拡張とbracingに関する実験を行う.
}%

\begin{figure}[t]
  \centering
  \includegraphics[width=1.0\columnwidth]{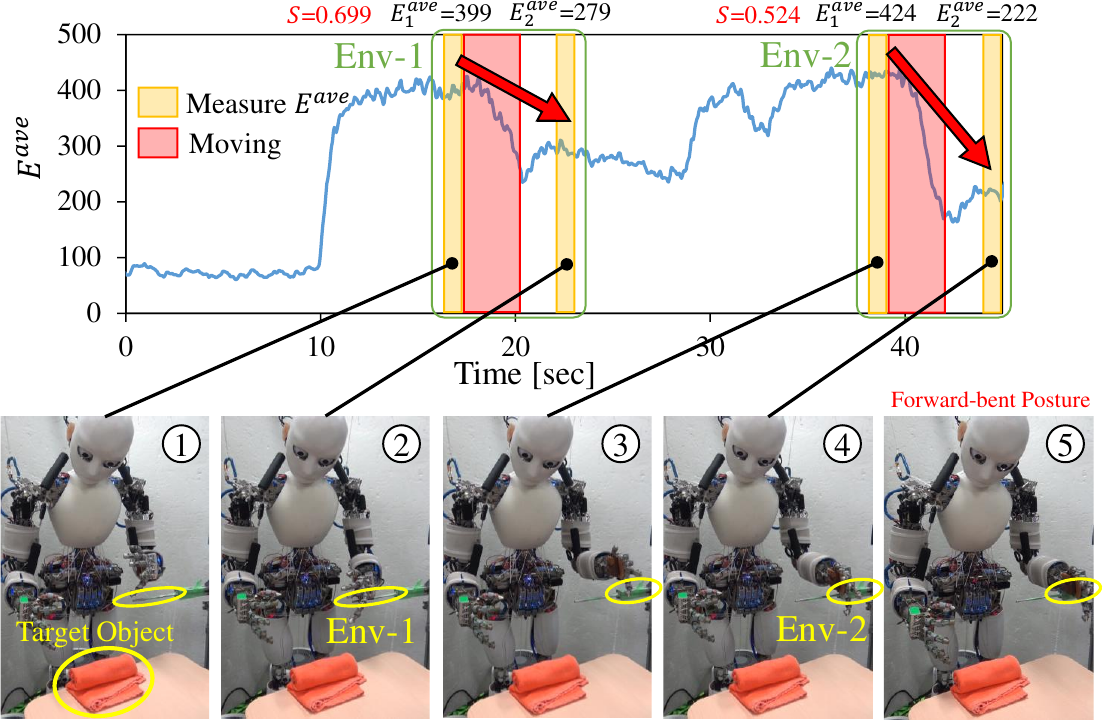}
  \vspace{-3.0ex}
  \caption{Motion extension: stability recognition when contacting Env-1 and Env-2.}
  \label{figure:extension-graph1}
  \vspace{-1.0ex}
\end{figure}

\begin{figure}[t]
  \centering
  \includegraphics[width=1.0\columnwidth]{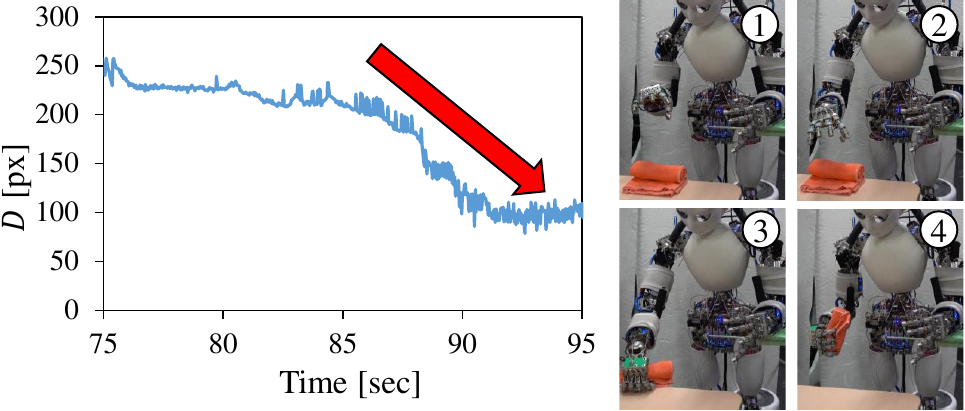}
  \vspace{-3.0ex}
  \caption{Motion extension: visual feedback with estimation of image Jacobian after stabilizing the body.}
  \label{figure:extension-graph2}
  \vspace{-3.0ex}
\end{figure}

\subsection{Motion Extension} \label{subsec:extension}
\switchlanguage%
{%
  To deal with a situation in which the robot posture is distorted when reaching for a target object, the robot stabilizes the self-body by attaching the hand to either Env-1 or Env-2, then leans forward, and reaches the target.
  Env-2 is a completely fixed and stable environment, whereas Env-1 is a thin and soft metal rod, which is an unstable environment.
  If the robot leans forward to Env-1, it will lose its balance and fall down because the rod is easily broken.
  The hand contacts the environment Env-1 and Env-2 in this order, the threshold of the stability evaluation value $S$ is set to 0.6, and the body is considered to be in a stable state if the value is smaller than the threshold.
  When measuring $E^{ave}$, we take the average of $E$ for 1 second, locally move for 3 seconds, wait for 2 seconds, and measure $E$ again for 1 second.
  For object grasping, we use the visual feedback method \cite{hosoda1994jacobian} that does not require the correct visual Jacobian.
  This is an important factor for the musculoskeletal humanoid which has a flexible body and is difficult to obtain the correct visual Jacobian.
  The stability recognition when contacting the environment is shown in \figref{figure:extension-graph1}, and the visual feedback after leaning forward to the environment is shown in \figref{figure:extension-graph2}.
  $S=0.699$ for Env-1 and $S=0.524$ for Env-2, indicating that the robot is able to maintain a stable posture after leaning forward to Env-2.
  Also, the distance $D$ between the pixels of the target object and the hand in the image gradually decreases with visual feedback, and the robot finally succeeds in grasping the object.
  Therefore, it is shown that by recognizing the stability, the robot can choose the environment in which to lean forward and stabilize the body to perform the task.
}%
{%
  ターゲットとなる物体を把持する際, そのまま手を伸ばすと姿勢が崩れてしまう状況に対して, 用意したEnv-1またはEnv-2に手をつけることで身体を安定化させた後に前傾して手を伸ばす実験を行う.
  Env-2は完全に固定され安定した環境であるのに対して, Env-1は細く柔らかい金属の棒であり, 不安定な環境である.
  Env-1に身体を預けると, 棒自体が細く折れやすいため, バランスを崩して倒れてしまう.
  Env-1, Env-2の順に環境に触れ, 安定性評価値$S$の閾値を0.6に設定し, それよりも小さな値であれば安定した環境であると見なして身体を預ける.
  $E^{ave}$を測定する際は1秒間の$E$の平均を取り, 3秒間で動作し, 2秒待ってまた$E$を一秒間計測する.
  物体把持の際は正しい視覚ヤコビアンを必要としないHosodaらの視覚フィードバック\cite{hosoda1994jacobian}を用いた.
  これは, 柔軟な身体を持ち正しい視覚ヤコビアンを得ることが難しい筋骨格ヒューマノイドに取って重要な要素となる.
  環境に接触する際の安定性認識を\figref{figure:extension-graph1}に, 前傾後の視覚フィードバックを\figref{figure:extension-graph2}に示す.
  Env-1のときは$S=0.699$, Env-2のときは$S=0.524$となり, Env-2に身体を預け前傾した後も姿勢を安定に保つことができていることがわかる.
  また, 視覚フィードバックにより徐々に画像中の手と物体の画素の距離$D$が減少し, 最終的に物体を把持することに成功している.
  よって, 安定性を認識することで身体を預ける環境を選び, 身体を安定化させてタスクを実行できることが示された.
}%

\begin{figure}[t]
  \centering
  \includegraphics[width=1.0\columnwidth]{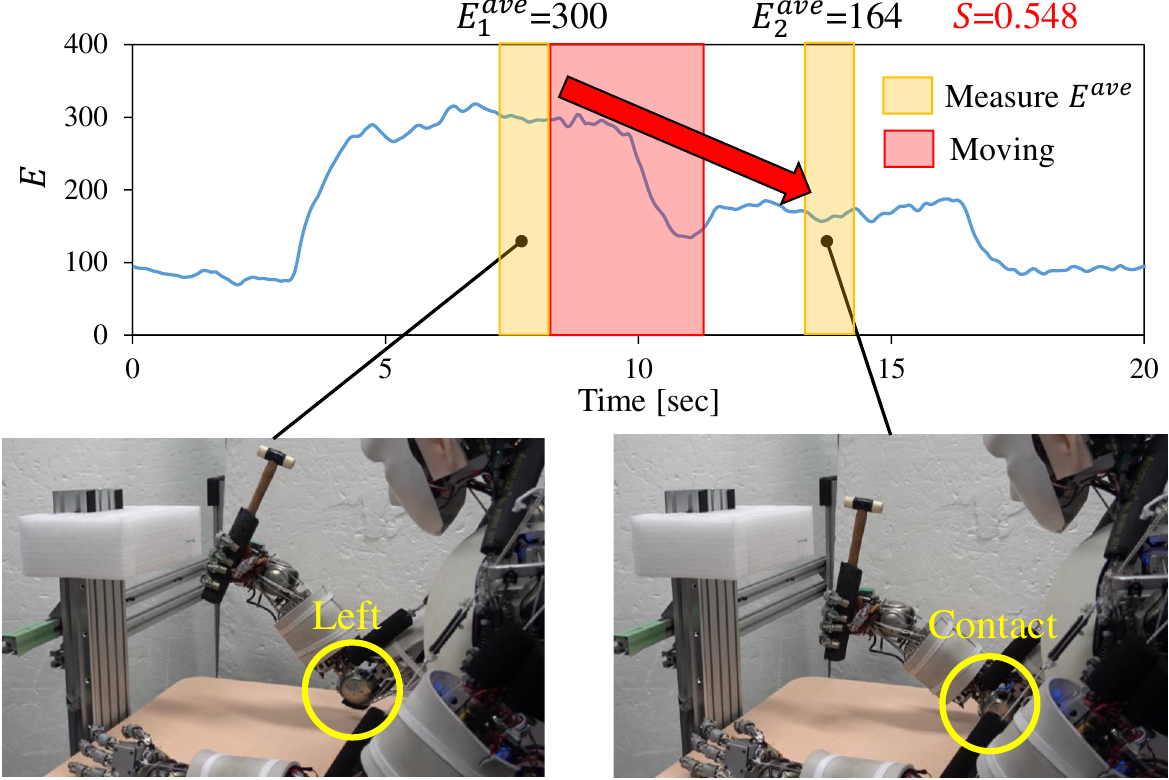}
  \vspace{-3.0ex}
  \caption{Bracing: stability recognition when contacting the table. The graph shows the transition of $E$.}
  \label{figure:bracing-graph1}
  \vspace{-1.0ex}
\end{figure}

\begin{figure}[t]
  \centering
  \includegraphics[width=1.0\columnwidth]{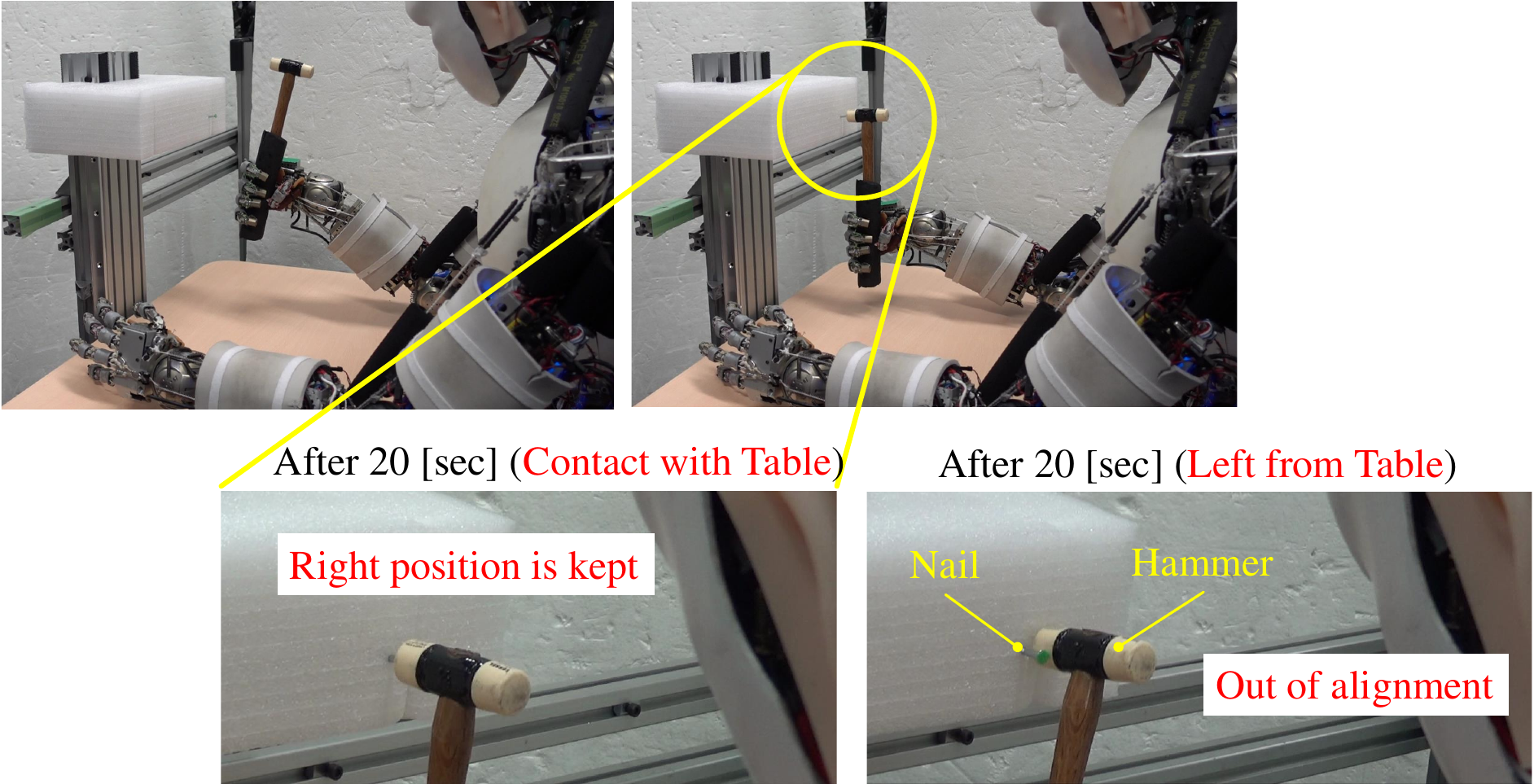}
  \vspace{-3.0ex}
  \caption{Bracing: hammer hitting after bracing. After 20 seconds of the experiment, the right position between a hammer and nail is kept with bracing, and is not kept without bracing.}
  \label{figure:bracing-graph2}
  \vspace{-3.0ex}
\end{figure}

\subsection{Bracing Behavior} \label{subsec:bracing}
\switchlanguage%
{%
  The robot puts its elbows on the table and hits a nail with a hammer precisely.
  We check how the stability of the robot changes depending on whether or not the robot puts its elbows on the table and how the hammering behavior changes.
  The evaluation value of the stability when the elbows are placed on the desk is shown in \figref{figure:bracing-graph1}, and the behavior of the hammer hitting with or without the elbows attached is shown in \figref{figure:bracing-graph2}.
  We can see that putting the elbow on the table changes $S$ and stabilizes the body.
  In addition, when the hammer is moved with the elbows on the desk, the positional relationship between the nail and hammer does not change even after 20 seconds of the hammer hitting, whereas it gradually changes in 20 seconds without the elbows on the desk.
  Therefore, the stabilization of the body by bracing can be estimated by $S$, and it is shown to be useful for accurate movements.
}%
{%
  机に肘をつき, 正確にハンマーを使って釘を打つ実験を行う.
  このとき, 肘をつけるかどうかによって安定性が変化し, ハンマーを打つ挙動がどう変化するかを確認する.
  机に肘を着いた際の安定性の評価を\figref{figure:bracing-graph1}に, 机につけた状態とつけない状態におけるハンマー動作の挙動を\figref{figure:bracing-graph2}に示す.
  肘を机につけることで, $S$が変化し, 身体を安定化させることができていることがわかる.
  また, 肘をつけてハンマーを動かした場合は20秒間同じ動作をしても常に釘とハンマーの位置関係が変化しなかったのに対して, 肘をつけない場合は徐々に位置関係が変化し, 20秒後にはハンマーの位置が釘からズレてしまった.
  よって, bracingによる身体の安定化が$S$により推定でき, 正確な動作にこれが有用であることが示された.
}%

\section{CONCLUSION} \label{sec:conclusion}
\switchlanguage%
{%
  In this study, we developed a method for recognizing the stability of the self-body by vibrating a part of the body and measuring the degree of propagation of the vibration from the sensory organ of a part of the body.
  For flexible musculoskeletal humanoids, the stability can be estimated by using the vibration in the muscle length command and the spectrum of the Fourier transform of the vibration in muscle tension.
  The characteristics of this estimator are clarified in terms of amplitude, frequency, grouping of muscles for vibration and observation, and differences in environment, and an appropriate use is proposed.
  Experiments on motion extension and bracing behavior using this estimator are conducted, and its effectiveness is confirmed.
  The important concept of this study, that is, the stability recognition with self vibration and its propagation, has been shown.

  In the future, we will apply this concept to various robots with more detailed verification from the theoretical point of view.
  The use of sensors such as accelerometers and contact sensors as well as muscle length and tension sensors is considered to be necessary.
  In addition, we would like to explore the ways in which the robots themselves can find and utilize the law of this study through their experiences.
}%
{%
  本研究では, 自己身体の一部を振動させ, その伝播度合いを身体の一部の感覚器から測定することで, 環境, そして自己の安定性を認識する手法を開発した.
  柔軟な筋骨格ヒューマノイドにおいて, 筋長力変化による振動と筋張力変化をフーリエ変換した際のスペクトルを使い, 安定性を推定することができる.
  振幅・周波数・振動と観測のグルーピング・環境の違いによる本推定器の特性を明らかにし, 適切な使い方を提案した.
  本推定器を用いて, 環境を使った動作拡張とbracingの実験を行い, その有効性を確認した.
  本研究の自身の振動とその変化による認識という重要なコンセプトは示すことができたと考える.

  今後理論面からのより詳細な検証を行い, 様々なロボットに本コンセプトを適用していきたい.
  筋長・筋張力センサだけでなく, 加速度センサや接触センサ等のセンサ利用が重要となると考えられる.
  また, 本ロボットのようなモデル化が難しいロボットにおいて, これら人間が見言い出した特性に関する知識をロボット自身が経験から獲得し, 利用していく方法について模索していきたい.
}%

\section*{Acknowledgement}
This research was supported by JST ACT-X Grant Number JPMJAX20A5 and JSPS KAKENHI Grant Number JP19J21672.
The authors would like to thank Yuka Moriya for proofreading this manuscript.

{
  \bibliographystyle{IEEEtran}
  \bibliography{main}
}

\end{document}